\pdfoutput=1
\documentclass[11pt]{article}

\usepackage{acl}

\usepackage{times}
\usepackage{latexsym}

\usepackage{booktabs}
\usepackage{multirow}

\usepackage{graphicx}

\usepackage[T1]{fontenc}
\usepackage[utf8]{inputenc}

\usepackage{microtype}

\title{The Power of Prompt Tuning for Low-Resource Semantic Parsing}

\usepackage{todonotes}
\makeatletter
\newcommand*\iftodonotes{\if@todonotes@disabled\expandafter\@secondoftwo\else\expandafter\@firstoftwo\fi}  % defines \iftodonotes{<true>}{<false>}, thanks to https://tex.stackexchange.com/questions/126559/conditional-based-on-packageoption
\makeatother

\author{
    Nathan Schucher\textsuperscript{1,2} ~
    Siva Reddy\textsuperscript{2,3} ~
    Harm de Vries\textsuperscript{1} ~ \\
    \textsuperscript{1}ServiceNow Research\\
    \textsuperscript{2}Mila/McGill University \\
    \textsuperscript{3}Facebook CIFAR AI Chair \\
    \texttt{\{nathan.schucher,harm.devries\}@servicenow.com}%
}

\begin{document}
\maketitle
\begin{abstract}
Prompt tuning has recently emerged as an effective method for adapting pre-trained language models to a number of language understanding and generation tasks. In this paper, we investigate prompt tuning for semantic parsing---the task of mapping natural language utterances onto formal meaning representations. On the low-resource splits of Overnight and TOPv2, we find that a prompt tuned  T5-xl significantly outperforms its fine-tuned counterpart, as well as strong GPT-3 and BART baselines. We also conduct ablation studies across different model scales and target representations, finding that, with increasing model scale,  prompt tuned T5 models improve at generating target representations that are far from the pre-training distribution. 
\end{abstract}

\section{Introduction}
With the widespread success of pre-trained language models (LMs;~\citealt{devlin2018pretraining,2020t5,stanford2021foundation}), it becomes increasingly important to explore how such models can be adapted to downstream tasks. One adaptation method which has recently attracted much attention is prompt design~\cite{brownLanguageModelsAre2020a,shin-etal-2020-autoprompt}, which modulates the behaviour of a LM through a task description and a few input-output examples. \citet{brownLanguageModelsAre2020a} show that this adaptation strategy is increasingly effective for larger LMs. However, prompt design is sensitive to the exact phrasing of the prompt, and, more importantly, performs worse than fine-tuning models on task-specific examples~\cite{lesterPowerScaleParameterEfficient2021}.

\begin{figure}[t]
\centering
\includegraphics[width=0.5\textwidth]{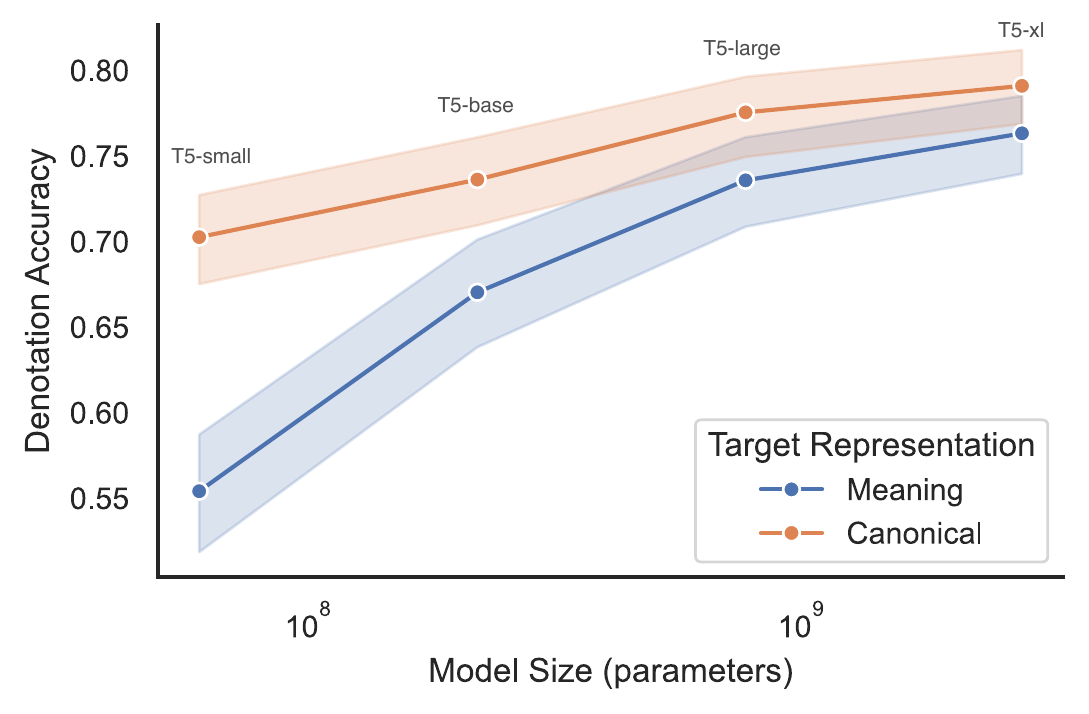}
\caption{We show that the T5 prompt tuning performance difference between target representations shrinks as the number of parameters increase, with constrained decoded T5-xl achieving close to performance parity.}
\label{fig:overnight_target_representation_gap}
\end{figure}

Prompt tuning has recently arisen as a strong performing alternative adaption method~\cite{lesterPowerScaleParameterEfficient2021}. Rather than hand-designing discrete prompts, prompt tuning optimizes the embeddings of a number of task-specific prompt tokens. In contrast to fine-tuning, this method keeps almost all LM parameters frozen. On a set of language understanding tasks, \citet{lesterPowerScaleParameterEfficient2021} show that prompt tuning becomes competitive with fine-tuning for the largest pre-trained T5 models~\cite{2020t5}. \citet{liPrefixTuningOptimizingContinuous2021} also explore a related parameter-efficient adaptation method called prefix-tuning, finding that it outperforms fine-tuning on low-resource natural language generation tasks. 

In this paper, we investigate prompt tuning for semantic parsing. This task is fundamentally different from the aforementioned language understanding and generation tasks, as it requires that models output formal meaning representations which do not resemble the natural language distribution seen during pre-training. In particular, we focus on the low-resource setup because examples for semantic parsing are difficult and expensive to collect~\cite{wangBuildingSemanticParser2015, marzoev2020unnatural}. We therefore evaluate prompt tuning on two datasets: the 200-shot version of Overnight~\cite{wangBuildingSemanticParser2015,shinConstrainedLanguageModels2021} and the low-resource splits TOPv2~\cite{chenLowResourceDomainAdaptation2020}. On both datasets, we compare prompt tuning T5 against fine-tuning and investigate the effect of canonicalizing the meaning representation, i.e. to what extent naturalizing the logical forms influences performance. In addition, we study the effect of T5 model scale on Overnight as well as varying data regimes on TOPv2. Our main findings can be summarized as follows:

\begin{itemize}
    \item For large T5 models, prompt tuning significantly outperforms fine-tuning in the low-data regime, resulting in an absolute improvement of 6\% and 15\% on Overnight and TOPv2, respectively. This performance gap decreases when more training data becomes available. 
    \item With growing model size, prompt tuned T5 models are increasingly capable of outputting diverse target representations (see Figure \ref{fig:overnight_target_representation_gap}). On Overnight, we find that the disparity between canonical and meaning representations shrinks from 17\% to 4\% for T5-small and T5-xl, respectively. On TOPv2, prompt tuned T5-large models are much better at generating out-of-vocabulary tokens than T5-small. 
\end{itemize}

\section{Related work}
Our work is related to recent work on semantic parsing and prompt tuning, which we briefly describe below. 

\subsection{Semantic Parsing}
Semantic parsing is the task of converting a natural language utterance $\mathbf{u} = (u_1,\dots, u_N)$ to a formal meaning representation $\mathbf{z} = (z_1,\dots, z_M)$. These meaning representations, also referred to as logical forms, can be interpreted by machines and executed in a real environment. For example, ThingTalk~\cite{campagna2019genie} and TOP~\cite{gupta-etal-2018-semantic-parsing} are meaning representations for executing commands of virtual assistants, while SQL is a representation for interacting with relational databases. In recent years, neural sequence-to-sequence models have become the dominant approach for semantic parsing tasks~\cite{dong-lapata-2016-language}.

\begin{figure*}[h]
\centering
\includegraphics[width=\textwidth]{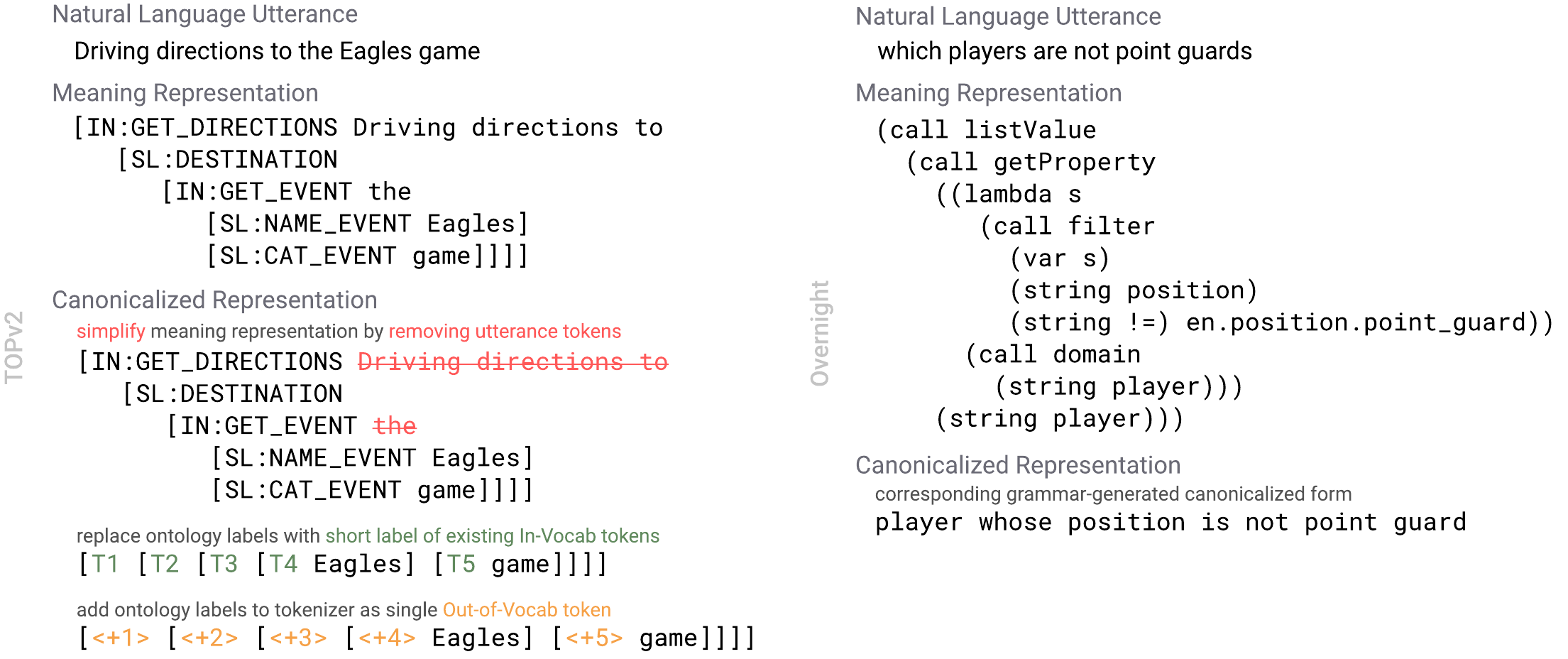}
\caption{Examples from the TOPv2 and Overnight datasets with the corresponding canonicalization schemes.}
\label{fig:canonicalized-examples}
\end{figure*}

\paragraph{Canonicalization} \label{sec:background-canonicalization} A common simplification step in semantic parsing is to canonicalize the meaning representations. That is, the meaning representation $\mathbf{z}$ is naturalized to a canonical form $\mathbf{c}$ through a grammar or set of rules. Examples of the meaning and canonical representation for Overnight and TOPv2 ~\cite{wangBuildingSemanticParser2015, chenLowResourceDomainAdaptation2020} can be found in Fig.~\ref{fig:canonicalized-examples}.

When canonical representations are available, \citet{berant-liang-2014-semantic} argue that semantic parsing can be seen as a paraphrase task. They propose to use a paraphrase model---using e.g. word vectors trained on Wikipedia---to find the best paraphrase of utterance $\mathbf{u}$ among a set of canonical utterances. They show this paraphrase model improves results over directly generating logical forms on two question-answering datasets. \citet{marzoev2020unnatural} extends this work by showing that pre-trained language models like BERT can be effective paraphrasers. While \citet{berant-liang-2014-semantic,marzoev2020unnatural} use  models to score canonical utterances, \citet{shinConstrainedLanguageModels2021} propose to constrain the generation process of autoregressive models like BART and GPT-3. On a number of few-shot semantic parsing tasks, they demonstrate the benefit of generating canonical representations over meaning representations. 

\subsection{Prompt-tuning}
\citet{lesterPowerScaleParameterEfficient2021} evaluates prompt tuning on SuperGLUE, a benchmark consisting of eight language understanding tasks. They find that prompt tuning becomes competitive with fine-tuning for the largest T5 model. \citet{liPrefixTuningOptimizingContinuous2021} propose prefix-tuning to adapt BART and GPT-2 for natural language generation tasks. This method differs from \citet{lesterPowerScaleParameterEfficient2021} in that it prepends trainable embeddings for each layer of the language model rather than  introducing token embeddings at the input layer. They demonstrate that pre-fix outperforms fine-tuning baselines. Similarly, \citet{liu2021gpt} also show encouraging results for prompt tuning on natural language understand and generation tasks. \citet{qin2021learning} also explores prompt tuning but for a knowledge extraction task. Inserting general adapter layers into pre-trained language models is also proposed in \citet{houlsby2019parameter,mahabadi2021compacter}.  Related to our work are also other few-shot adaptation techniques like PET~\cite{schickExploitingClozeQuestionsFewShot2021a}.  Moreover, adapter layers have also been explored in the computer vision domain ~\cite{Rebuffi17,de2017modulating}. 

\section{Experiments}
To evaluate low-resource prompt tuning, we compare against fine-tuned variants of the same model on two semantic parsing datasets with canonical representations available. We compare both large and small variants of the T5 architecture on these datasets and experiment with various canonicalized representations.

\subsection{Datasets} \label{sec:exp-datasets}

\paragraph{Overnight} The Overnight semantic parsing dataset~\cite{wangBuildingSemanticParser2015} consists of 13,682 natural utterance, canonical form, meaning representation triples split across eight domains.
% The task is to map a natural language utterance to the corresponding meaning representation. The dataset was collected by first enumerating all pairs of canonical utterances and meaning representations from a synchronous context-free grammar to a fixed depth. These human-interpretable, but unnatural utterances, were subsequently paraphrased by crowd-sourced workers into natural language utterances. 
To simulate low-resource splits of this dataset, we follow \citeauthor{shinConstrainedLanguageModels2021} and create randomly subsampled splits of 200 training examples for each domain, using 20\% of the remaining data for validation. We measure and report denotation accuracy by evaluating all predicted queries using the SEMPRE toolkit \cite{berant2013freebase}. We repeat each experiment on Overnight with five different random splits.

\paragraph{TOPv2}
\citet{chenLowResourceDomainAdaptation2020} introduce the TOPv2 dataset, a task-oriented semantic parsing dataset with eight domains, two of which come with pre-defined low-resource splits. The authors propose a principled way of constructing low-resource training sets, \textit{samples per intent and slot} (SPIS), intended to ensure equal exposure to ontology labels across domains of varying complexity. We experiment with the \textit{weather} and \textit{reminder} domains at the 10, 25, and 500 SPIS resource splits, performing five runs on each model varying the random seed. The \textit{reminder} domain is the most challenging with 19 intent labels, 32 slot labels, and with 21\% of the programs having a depth greater than 2. \textit{Weather} in comparison has 7 intent labels, 11 slot labels, and no programs with depth greater than 2. 

\subsection{Canonicalized Representations} \label{sec:exp-canonicalization}

\begin{table*}[h]
\scriptsize
\centering
\begin{tabular}{lllrrrrrrrrr}
\toprule
{\textbf{Model}} & {\textbf{Representation}} & {\textbf{Method}} & {\textbf{Bask.}} & {\textbf{Blo.}} & {\textbf{Cal.}} & {\textbf{Hou.}} & {\textbf{Pub.}} & {\textbf{Rec.}} & {\textbf{Res.}} & {\textbf{Soc.}} & {\textbf{Avg}} \\

\midrule
\multirow[t]{4}{*}{T5-small} & \multirow[t]{2}{*}{Meaning} & FT & 0.767 & 0.454 & 0.685 & 0.608 & 0.640 & 0.698 & 0.691 & 0.581 & 0.641 \\
 &  & PT & 0.621 & 0.312 & 0.470 & 0.352 & 0.478 & 0.506 & 0.608 & 0.352 & 0.463 \\
 & \multirow[t]{2}{*}{Canonical} & FT & 0.775 & 0.466 & 0.721 & 0.616 & 0.665 & 0.673 & 0.636 & 0.568 & 0.640 \\
 &  & PT & 0.764 & 0.440 & 0.680 & 0.601 & 0.648 & 0.699 & 0.697 & 0.578 & 0.638 \\
\midrule
\multirow[t]{4}{*}{T5-base} & \multirow[t]{2}{*}{Meaning} & FT & 0.769 & 0.455 & 0.717 & 0.612 & 0.670 & 0.713 & 0.714 & 0.587 & 0.655 \\
 &  & PT & 0.717 & 0.429 & 0.677 & 0.510 & 0.596 & 0.639 & 0.705 & 0.492 & 0.596 \\
 & \multirow[t]{2}{*}{Canonical} & FT & 0.800 & 0.466 & 0.736 & 0.642 & 0.711 & 0.694 & 0.696 & 0.597 & 0.668 \\
 &  & PT & 0.786 & 0.452 & 0.682 & 0.636 & 0.675 & 0.705 & 0.733 & 0.614 & 0.660 \\
\midrule
\multirow[t]{2}{*}{BART} & Meaning & FT & 0.734 & 0.370 & 0.514 & 0.540 & 0.514 & 0.477 & 0.417 & 0.424 & 0.499 \\
 & Canonical & FT & 0.591 & 0.331 & 0.740 & 0.309 & 0.668 & 0.598 & 0.582 & 0.532 & 0.544 \\
\midrule
\multirow[t]{4}{*}{T5-large} & \multirow[t]{2}{*}{Meaning} & FT & 0.777 & 0.432 & 0.690 & 0.639 & 0.709 & 0.729 & 0.723 & 0.590 & 0.661 \\
 &  & PT & 0.792 & 0.469 & 0.739 & 0.676 & 0.696 & 0.734 & 0.778 & 0.600 & 0.685 \\
 & \multirow[t]{2}{*}{Canonical} & FT & 0.793 & 0.458 & 0.760 & 0.658 & 0.678 & 0.727 & 0.715 & 0.581 & 0.671 \\
 &  & PT & 0.819 & 0.525 & 0.768 & 0.712 & 0.744 & 0.789 & 0.769 & 0.655 & 0.723 \\
\midrule
\multirow[t]{4}{*}{T5-xl} & \multirow[t]{2}{*}{Meaning} & FT & 0.774 & 0.413 & 0.702 & 0.630 & 0.682 & 0.691 & 0.705 & 0.580 & 0.647 \\
 &  & PT & 0.819 & 0.532 & 0.767 & 0.693 & 0.694 & 0.758 & 0.778 & 0.632 & 0.709 \\
 & \multirow[t]{2}{*}{Canonical} & FT & 0.799 & 0.486 & \bfseries 0.781 & 0.647 & 0.724 & 0.732 & 0.725 & 0.619 & 0.689 \\
 &  & PT & \bfseries 0.839 & \bfseries 0.544 & 0.777 & \bfseries 0.729 & \bfseries 0.770 & \bfseries 0.791 & \bfseries 0.789 & \bfseries 0.702 & \bfseries 0.743 \\
\bottomrule
\end{tabular}

\caption{Unconstrained denotation accuracy for all models (with unconstrained decoding) on the Overnight dataset. For each domain, we report the average over 5 runs trained on randomly sampled splits of 200 examples for fine-tuned (FT) and prompt tuned (PT) models. }
\label{table:overnight_canonicalization}
\end{table*}

\subsubsection{Overnight}
Overnight uses a context-free synchronous grammar to generate canonical representations for the logical forms. As can be seen in Fig.~\ref{fig:canonicalized-examples}, these canonical representations resemble natural language. 
%While the original paper used these synthetic utterances as templates for crowd-sourcing natural language utterances, subsequent work has used them as targets for semantic parsing models~\cite{marzoev2020unnatural,shinConstrainedLanguageModels2021}. We follow this work and treat these paraphrased utterances as our target canonicalized representations to compare with the normal Overnight meaning representations.

\subsubsection{TOPv2} \citeauthor{chenLowResourceDomainAdaptation2020} apply a set of simple modifications to the TOPv2 meaning representations to arrive at a canonical form used in all their experiments. Unlike Overnight, these pre-processing steps are largely small encoding differences and do not change the syntactic structure of the logical forms. We adopt all of these canonicalization steps (except for lexicographic sorting of the semantic parse tree) and add an ontology label shortening step. Examples of these transformations can be seen in Fig.~\ref{fig:canonicalized-examples} and are briefly described below. 

\begin{description}
    \item[Simplify] removes redundant utterance tokens unnecessary for interpreting the meaning representation.
    \item[Out-of-Vocab] adds the entire intent or slot label to the tokenizer as a new single tokens with a corresponding randomly initialized embedding.
    \item[In-Vocab] replaces the intent and slot labels with a short unique identifier representable by the pre-trained tokenizer.
\end{description}

We perform an ablation over these canonicalization choices, repeating each experiment three times with varying random seed.

\subsection{Models}
We provide training details and hyperparameters for all models in Appendix \ref{sec:exp-models}. Below, we briefly explain the prompt-tuning methodology. 

\subsubsection{Prompt Tuning}
Prompt tuning, as proposed by \citet{lesterPowerScaleParameterEfficient2021}, prepends a sequence of continuous embeddings $\mathbf{p} = (p_1, \dots, p_K)$ to the sequence input embeddings $e(\mathbf{u}) = (e(u_1),\dots, e(u_N))$ before feeding it to a language model with parameters $\mathbf{\theta}$. During prompt tuning we optimize the prompt embeddings $(p_1, \dots, p_K)$ exclusively, keeping the language model parameters $\mathbf{\theta}$ and the pretrained vocabulary embeddings fixed. Note that this process still requires backpropagating gradients through the full language model. Like fine-tuning models, we maximize the likelihood of generating the output sequence $\mathbf{z}$.

\begin{table*}[h]
\scriptsize
\centering
\begin{tabular}{llllrrrrrrrrr}
\toprule
{\textbf{Model}} & {\textbf{Representation}} & {\textbf{Method}} & {\textbf{Decoding}} & {\textbf{Bask.}} & {\textbf{Blo.}} & {\textbf{Cal.}} & {\textbf{Hou.}} & {\textbf{Pub.}} & {\textbf{Rec.}} & {\textbf{Res.}} & {\textbf{Soc.}} & {\textbf{Avg}} \\
\midrule
% T5-xl & Meaning & FT & Constrained & 0.799 & 0.453 & 0.731 & 0.658 & 0.749 & 0.724 & 0.728 & 0.647 & 0.686 \\
T5-xl & Meaning & PT & Constrained & 0.841 & 0.592 & 0.802 & 0.765 & 0.776 & 0.814 & 0.789 & 0.725 & 0.763 \\
% T5-xl & Canonical & FT & Constrained & 0.818 & 0.555 & 0.783 & 0.705 & 0.763 & 0.770 & 0.752 & 0.703 & 0.731 \\
T5-xl & Canonical & PT & Constrained & 0.856 & 0.619 & 0.806 & \bfseries 0.779 & \bfseries 0.824 & 0.830 & 0.822 & \bfseries 0.793 & \bfseries 0.791 \\
\midrule
BART$^\dagger$ & Meaning & FT & Constrained & 0.834 & 0.499 & 0.750 & 0.619 & 0.739 & 0.796 & 0.774 & 0.620 & 0.704 \\
BART$^\dagger$ & Canonical & FT & Constrained & \bfseries 0.864 & 0.554 & 0.780 & 0.672 & 0.758 & 0.801 & 0.801 & 0.666 & 0.737 \\
\midrule
GPT-2$^\dagger$  & Meaning & FT & Constrained & 0.760 & 0.479 & 0.736 & 0.571 & 0.645 & 0.699 & 0.660 & 0.606 & 0.644 \\
GPT-2$^\dagger$ & Canonical & FT & Constrained & 0.836 & 0.540 & 0.766 & 0.666 & 0.715 & 0.764 & 0.768 & 0.623 & 0.710 \\
\midrule
GPT-3$^\dagger$ & Canonical & Context & Constrained & 0.859 & \bfseries 0.634 & 0.792 & 0.741 & 0.776 & 0.792 & 0.840 & 0.687 & 0.765 \\
\midrule
GPT-3$^\dagger$$^*$ & Meaning & Context & Constrained & 0.680 & 0.530 & 0.680 & 0.580 & 0.630 & 0.750 & 0.780 & 0.630 & 0.657 \\
GPT-3$^\dagger$$^*$ & Canonical & Context & Constrained & 0.800 & 0.620 & \bfseries 0.820 & 0.710 & 0.790 & \bfseries 0.840 & \bfseries 0.890 & 0.720 & 0.774 \\

\bottomrule
\end{tabular}

\caption{Constrained denotation accuracy for all models on the Overnight dataset. For each domain, we report the average over 5 runs trained on randomly sampled splits of 200 examples. $^\dagger$ denotes results reported by \citet{shinConstrainedLanguageModels2021}. $^*$ indicates performance on subsampled test set.}
\label{table:overnight_constrained}
\end{table*}

\section{Results}
In Table \ref{table:overnight_canonicalization}, we report Overnight results across four T5 model scales and two target representations. In Table \ref{table:overnight_constrained}, we add constrained decoding (see Appendix \ref{sec:exp-models}) to our best performing T5 model and compare against previously reported Overnight results. In Table \ref{table:topv2_results}, we display the results of T5-large on the three different SPIS-splits of TOPv2, and include the BART-CopyPtr results from \citet{chenLowResourceDomainAdaptation2020}. In Table \ref{table:topv2_canonicalization}, we summarize the results of the canonicalization ablation study for TOPv2.

\subsection{Prompt tuning vs fine tuning}

\begin{table}[h]
\scriptsize
\centering
\begin{tabular}{lllrrr}
\toprule
{\textbf{SPIS}} & {\textbf{Model}} & {\textbf{Method}} & {\textbf{Reminder}} & {\textbf{Weather}} & {\textbf{Average}} \\
\midrule
\multirow[t]{2}{*}{10} & \multirow[t]{2}{*}{T5-large} & FT & 0.392 & 0.579 & 0.486 \\
 &  & PT & \bfseries 0.567 & \bfseries 0.700 & \bfseries 0.634 \\
\midrule\multirow[t]{3}{*}{25} & BART-CopyPtr & FT & 0.557 & 0.716 & 0.637 \\
 & \multirow[t]{2}{*}{T5-large} & FT & 0.502 & 0.683 & 0.593 \\
 &  & PT & \bfseries 0.642 & \bfseries 0.739 & \bfseries 0.691 \\
\midrule\multirow[t]{3}{*}{500} & BART-CopyPtr & FT & 0.719 & \bfseries 0.849 & 0.784 \\
 & \multirow[t]{2}{*}{T5-large} & FT & 0.649 & 0.846 & 0.748 \\
 &  & PT & \bfseries 0.749 & 0.847 & \bfseries 0.798 \\
\bottomrule
\end{tabular}

\caption{Average exact match accuracies (5 runs) for different low-resource splits of the TOPv2 dataset. BART-CopyPtr results from \citet{chenLowResourceDomainAdaptation2020}.}
\label{table:topv2_results}
\end{table}

We find that prompt tuning improves over fine-tuning for all large model configurations and target representations. On Overnight, prompt tuned denotation accuracy exceeds fine-tuned counterparts by up to 5 points with T5-large and T5-xl. For T5-small and T5-base, prompt tuning remains competitive (within 1\% average accuracy) with fine-tuning when predicting canonical forms. On TOPv2, prompt tuning achieves an absolute improvement of 15\% mean accuracy over fine-tuning on the lowest SPIS split. This performance disparity lessens when training data increases; however, prompt tuned T5-large continues to beat its fine-tuned counterpart by 5 points at 500 SPIS and the BART-CopyPtr model by 1.4 points. 

Our prompt tuning models outperform previously reported results on these datasets. On Overnight, our best model---T5-xl PT with canonical representations and constrained decoding---outperforms the BART FT model of \citet{shinConstrainedLanguageModels2021} by 5 accuracy points, and GPT-3 by more than 2 points. On the 25 SPIS split of TOPv2, we see an average improvement of more than 5 points compared to the BART-CopyPTR of \citet{chenLowResourceDomainAdaptation2020}.

\subsection{Canonical vs meaning representations}
Our main finding is that prompt tuned T5 models become better at generating meaning representations with increased model size. On Overnight, we see the absolute difference between canonical and meaning representations shrink from 17.5 points for T5-small to 3.4 points for T5-xl (Table \ref{table:overnight_canonicalization}). This gap shrinks another 18\% to 2.8 points when we apply constrained decoding to T5-xl (Table \ref{table:overnight_constrained}). By contrast, \citet{shinConstrainedLanguageModels2021} reports an 11.7 point difference when prompting GPT-3. For our fine-tuning baselines, we observe a small performance gap of 4 points across target representations for BART and T5-xl, while we observe no gap for T5-small, T5-base, and T5-large models.

% For T5-large, we find very small gains for generating canonical representations instead of meaning representations. On overnight (Table \ref{table:overnight_canonicalization}), the gap is 0.4 accuracy points for T5-large. This difference is much smaller than previously reported for unconstrained decoding with BART and GPT-3 by \citet{shinConstrainedLanguageModels2021} (3 and 13 points, respectively).  Interestingly, we do not find such a gap for canonical and meaning representations when fine-tuning T5 models, indicating there are qualitative differences between BART and T5 models.  

%On TOPv2, we find a similar pattern that large T5 models can extrapolate far beyond the training distribution. We especially observe this when adding novel vocabulary tokens (\verb|Out-of-Vocab| in Table \ref{table:topv2_canonicalization}), arguably the most out-of-distribution canonicalization choice. These novel token embeddings are randomly initialized and frozen during prompt tuning. Nevertheless, a prompt tuned T5-large model can successfully generate such sequences whereas a prompt tuned T5-small model struggles. Specifically, T5-large only sees a modest decrease from 70\% to 68.9\% on 10 SPIs and from 84.7.9\% to 83.3\% on the 500 SPIS (see \verb|None| and \verb|Vocab| columns in Table \ref{table:topv2_canonicalization}, respectively). In contrast, T5-small drops from 43.4\% to 23.4\% on 10 SPIs and from 71.8\% to 48.3\% on 500 SPIs. 

In our TOPv2 experiments we find similar evidence of large T5 model flexibility for generating sequences far from the training distribution. In particular, for our most intrusive canonicalization scheme \verb|Out-of-Vocab|, which adds novel tokens to the vocabulary and leaves these embeddings un-trained, we find no significant reduction in performance for T5-large across all data resource levels. T5-small, in comparison, sees almost a 50\% drop in performance relative to no canonicalization (\verb|None|) at the 10 SPIS level and continues to underperform by 33 \% at the 500 SPIS level.

Interestingly, we find that \verb|In-Vocab| drastically reduces performance for T5-small at the 10 SPIS level---30.9\% vs. 43.4\% for \verb|None|---but slightly outperforms it at 500 SPIS. We speculate that \verb|In-Vocab| effectively anonymizes the ontology tokens, obscuring information that is useful for prediction. In low-data regimes there is not enough training data to learn the semantics of these anonymized tokens, whereas with enough data this problem vanishes. 

\begin{table}[t]
\scriptsize
\centering
\begin{tabular}{lllllllll}
\toprule

{} & \multicolumn{2}{c}{\textbf{None}} & \multicolumn{2}{c}{\textbf{Simplified}} & \multicolumn{2}{c}{\textbf{In-Vocab}} & \multicolumn{2}{c}{\textbf{Out-of-Vocab}} \\
\cmidrule(lr){2-3} \cmidrule(lr){4-5}  \cmidrule(lr){6-7} \cmidrule(lr){8-9} 

{\textbf{SPIS}} & {Sm.} & {Lg.} & {Sm.} & {Lg.} & {Sm.} & {Lg.} & {Sm.} & {Lg.} \\
\midrule
10 & 0.43 & \bfseries 0.70 & 0.31 & 0.66 & 0.45 & 0.64 & 0.23 & 0.69 \\
25 & 0.56 & \bfseries 0.74 & 0.51 & 0.73 & 0.55 & 0.71 & 0.27 & 0.70 \\
500 & 0.72 & \bfseries 0.85 & 0.72 & \bfseries 0.85 & 0.72 & \bfseries 0.85 & 0.48 & 0.83 \\
\bottomrule
\end{tabular}
\caption{Exact match accuracies (3 runs) on TOPv2 Weather domain for different meaning representation canonicalization choices (\textbf{bold} indicates best exact match accuracy at that resource level), Sm. and Lg. refer to T5-small and T5-large, respectively.}
\label{table:topv2_canonicalization}
\end{table}

\section{Conclusion}
We find that prompt tuning is an effective method for adapting language models to the semantic parsing task. Prompt tuning significantly outperforms fine-tuning in low-data regimes, and remains competitive in the fully supervised setting.  We furthermore find that while canonicalizing meaning representations can slightly improve performance, the disparity between target representations decreases when prompt tuning larger T5 models. This result differs from previous work~\cite{shinConstrainedLanguageModels2021} which suggested that pre-trained LMs are much better equipped to output canonical than meaning representations. However, a significant limitation of prompt tuning is that it takes more time to converge than fine-tuning. We believe one fruitful direction for future research is to find ways to reduce the compute required to prompt tune.

\clearpage
\section{Ethical Considerations and Limitations}

There are two main limitations of this work. The first is the limited analysis of the learned prompts. While concurrent work has shown that interpreting prompts is a difficult task, it is still an important consideration and left for future work \cite{khashabiPROMPTWAYWARDNESSCurious2021}. Secondly, training prompts on meaning representations requires substantially more compute than fine-tuning. This may exacerbate inequalities in regions where access to data and compute are similarly limited \cite{ahia2021low}.

% Entries for the entire Anthology, followed by custom entries
\bibliography{anthology,custom}
\bibliographystyle{acl_natbib}

% This is an appendix.

\appendix
\clearpage

\section{Models} \label{sec:exp-models}
Here we provide all model details and hyperparameters to reproduce our results. We experiment with BART and T5 \cite{lewis-etal-2020-bart, 2020t5}, two large pre-trained encoder-decoder language models. BART is trained on the same 160GB text dataset used to train RoBERTa \cite{, lewis-etal-2020-bart} with a denoising objective. There are two size configurations (BART-base, BART-large) and we experiment only with the 406M parameter BART-large on the Overnight dataset. T5 is trained on the 750GB C4 dataset \cite{2020t5} with a de-noising objective. We use the T5-v1.1 checkpoints from \citet{lesterPowerScaleParameterEfficient2021} that were trained for an additional 100K steps with the Prefix-LM objective.
T5-v1.1 has five configurations at various scales: small, base, large, xl, xxl which have 60M, 220M, 770M, 3B, and 11B parameters, respectively. Here, we experiment with models up to T5-xl. All experiments were run with PyTorch (v. 1.8.1) and the Huggingface Transformers (v. 4.8.2) library \cite{pytorch-2019,wolf-etal-2020-transformers}.

\paragraph{Fine-tuning baseline}
We compare against baselines that fine-tune all parameters of BART and T5. We train the T5 models with AdaFactor \cite{Shazeer2018AdafactorAL} and BART with Adam \cite{lewis-etal-2020-bart, Kingma2015AdamAM}. On TOPv2, we use a learning rate of $10^{-4}$ and batch size of 128. On Overnight, we use a learning rate of $10^{-3}$ and a batch size of 64 across all sizes of T5. On both datasets, we train for 5000 epochs and perform model selection by early stopping on the validation set. 

\paragraph{Prompt tuning}
We follow the prompt tuning procedure proposed by \citeauthor{lesterPowerScaleParameterEfficient2021} for T5. We use 150 prompt tokens for all model sizes with a learning rate of 0.3 optimized with AdaFactor. We train for 5000 epochs on most domains, although it sometimes took as many as 20000 epochs to converge on the low-resource splits. Like the fine-tuned baseline, we perform model selection with best exact match accuracy on the validation set. We apply the same method to BART and found that it did not converge under a number of hyperparameter configurations. We therefore exclude prompt tuned BART models from our results\footnote{\citeauthor{liPrefixTuningOptimizingContinuous2021} also find that prompt tuning with BART is unstable and parameterize the prefix with an MLP; we did not attempt this setup.}.

\paragraph{Constrained Decoding}
We implement grammar-constrained decoding by building a prefix tree containing all canonical or meaning representations in the dataset as in \citet{shinConstrainedLanguageModels2021}. When doing constrained decoding we perform a beam search with 10 beams and use the prefix tree to look up valid single token continuations of the decoded sequence.

\section{Results} \label{sec:appendix-results}
For completeness, we provide all Overnight results in Table \ref{table:all-overnight-results}. 

\subsection{Training Times}
\begin{figure}[h]
\centering
\includegraphics[width=0.5\textwidth]{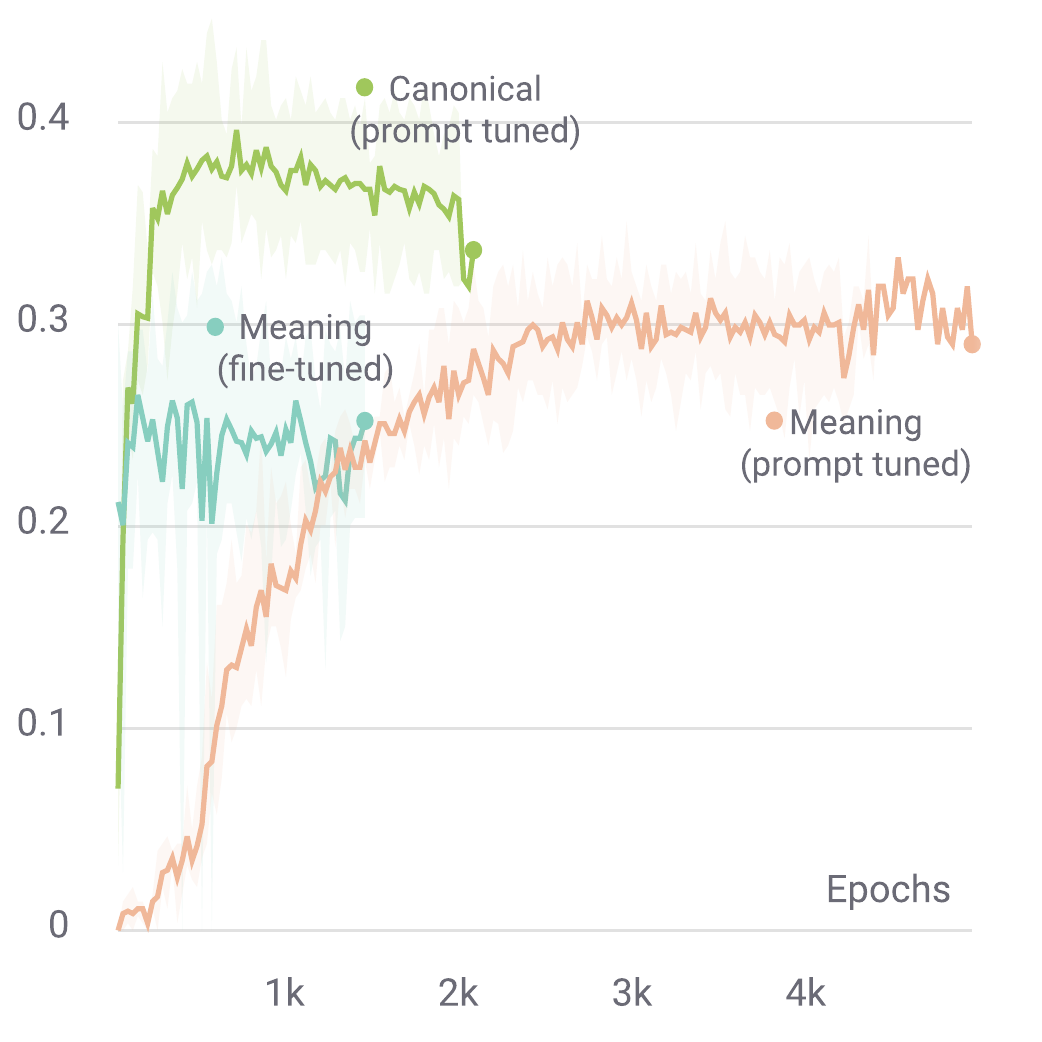}
\caption{Prompt and fine-tuned exact match validation accuracy on the Overnight \textit{blocks} domain. Fine-tuned models can quickly reach peak validation accuracy regardless of target representation. Prompt tuned models can take thousands of epochs to converge when predicting meaning representations.}
\label{fig:val-accuracy-canonical}
\end{figure}

Prompt tuned parameter efficiency comes at a cost: we find that prompt tuning takes significantly longer to train with early stopping than does fine-tuning. On the Overnight dataset, fine-tuned models typically took 250 epochs before validation performance plateaued. Our prompt tuned models frequently took more than 1000 epochs when predicting canonical representations, and up to 5,000 when predicting meaning representations. In Figure \ref{fig:val-accuracy-canonical}, we show example training curves for prompt tuning and fine-tuning. 

\begin{table*}[h]
\tiny
\centering
\begin{tabular}{llllrrrrrrrrr}
\toprule
{\textbf{Model}} & {\textbf{Representation}} & {\textbf{Method}} & {\textbf{Decoding}} & {\textbf{Bask.}} & {\textbf{Blo.}} & {\textbf{Cal.}} & {\textbf{Hou.}} & {\textbf{Pub.}} & {\textbf{Rec.}} & {\textbf{Res.}} & {\textbf{Soc.}} & {\textbf{Avg}} \\
\midrule
\multirow[t]{8}{*}{T5-small} & \multirow[t]{4}{*}{Meaning} & \multirow[t]{2}{*}{FT} & Unconstrained & 0.767 & 0.454 & 0.685 & 0.608 & 0.640 & 0.698 & 0.691 & 0.581 & 0.641 \\
 &  &  & Constrained & 0.787 & 0.519 & 0.725 & 0.624 & 0.753 & 0.752 & 0.705 & 0.664 & 0.691 \\
 &  & \multirow[t]{2}{*}{PT} & Unconstrained & 0.621 & 0.312 & 0.470 & 0.352 & 0.478 & 0.506 & 0.608 & 0.352 & 0.463 \\
 &  &  & Constrained & 0.656 & 0.392 & 0.615 & 0.475 & 0.593 & 0.588 & 0.663 & 0.450 & 0.554 \\
 & \multirow[t]{4}{*}{Canonical} & \multirow[t]{2}{*}{FT} & Unconstrained & 0.775 & 0.466 & 0.721 & 0.616 & 0.665 & 0.673 & 0.636 & 0.568 & 0.640 \\
 &  &  & Constrained & 0.811 & 0.519 & 0.744 & 0.663 & 0.729 & 0.723 & 0.692 & 0.671 & 0.694 \\
 &  & \multirow[t]{2}{*}{PT} & Unconstrained & 0.764 & 0.440 & 0.680 & 0.601 & 0.648 & 0.699 & 0.697 & 0.578 & 0.638 \\
 &  &  & Constrained & 0.787 & 0.521 & 0.730 & 0.679 & 0.735 & 0.748 & 0.746 & 0.674 & 0.703 \\
\midrule
\multirow[t]{8}{*}{T5-base} & \multirow[t]{4}{*}{Meaning} & \multirow[t]{2}{*}{FT} & Unconstrained & 0.769 & 0.455 & 0.717 & 0.612 & 0.670 & 0.713 & 0.714 & 0.587 & 0.655 \\
 &  &  & Constrained & 0.790 & 0.496 & 0.738 & 0.639 & 0.743 & 0.745 & 0.737 & 0.644 & 0.692 \\
 &  & \multirow[t]{2}{*}{PT} & Unconstrained & 0.717 & 0.429 & 0.677 & 0.510 & 0.596 & 0.639 & 0.705 & 0.492 & 0.596 \\
 &  &  & Constrained & 0.754 & 0.494 & 0.760 & 0.593 & 0.725 & 0.699 & 0.752 & 0.586 & 0.670 \\
 & \multirow[t]{4}{*}{Canonical} & \multirow[t]{2}{*}{FT} & Unconstrained & 0.800 & 0.466 & 0.736 & 0.642 & 0.711 & 0.694 & 0.696 & 0.597 & 0.668 \\
 &  &  & Constrained & 0.840 & 0.525 & 0.745 & 0.676 & 0.773 & 0.736 & 0.734 & 0.696 & 0.716 \\
 &  & \multirow[t]{2}{*}{PT} & Unconstrained & 0.786 & 0.452 & 0.682 & 0.636 & 0.675 & 0.705 & 0.733 & 0.614 & 0.660 \\
 &  &  & Constrained & 0.826 & 0.550 & 0.774 & 0.717 & 0.780 & 0.764 & 0.770 & 0.708 & 0.736 \\
\midrule
\multirow[t]{2}{*}{BART} & Meaning & FT & Unconstrained & 0.734 & 0.370 & 0.514 & 0.540 & 0.514 & 0.477 & 0.417 & 0.424 & 0.499 \\
 & Canonical & FT & Unconstrained & 0.591 & 0.331 & 0.740 & 0.309 & 0.668 & 0.598 & 0.582 & 0.532 & 0.544 \\
\midrule
\multirow[t]{8}{*}{T5-large} & \multirow[t]{4}{*}{Meaning} & \multirow[t]{2}{*}{FT} & Unconstrained & 0.777 & 0.432 & 0.690 & 0.639 & 0.709 & 0.729 & 0.723 & 0.590 & 0.661 \\
 &  &  & Constrained & 0.789 & 0.475 & 0.713 & 0.662 & 0.743 & 0.754 & 0.717 & 0.641 & 0.687 \\
 &  & \multirow[t]{2}{*}{PT} & Unconstrained & 0.792 & 0.469 & 0.739 & 0.676 & 0.696 & 0.734 & 0.778 & 0.600 & 0.685 \\
 &  &  & Constrained & 0.816 & 0.533 & 0.774 & 0.742 & 0.760 & 0.787 & 0.793 & 0.680 & 0.736 \\
 & \multirow[t]{4}{*}{Canonical} & \multirow[t]{2}{*}{FT} & Unconstrained & 0.793 & 0.458 & 0.760 & 0.658 & 0.678 & 0.727 & 0.715 & 0.581 & 0.671 \\
 &  &  & Constrained & 0.819 & 0.509 & 0.751 & 0.703 & 0.718 & 0.742 & 0.728 & 0.664 & 0.704 \\
 &  & \multirow[t]{2}{*}{PT} & Unconstrained & 0.819 & 0.525 & 0.768 & 0.712 & 0.744 & 0.789 & 0.769 & 0.655 & 0.723 \\
 &  &  & Constrained & 0.841 & 0.597 & 0.805 & 0.770 & 0.794 & 0.823 & 0.823 & 0.750 & 0.775 \\
\midrule
\multirow[t]{8}{*}{T5-xl} & \multirow[t]{4}{*}{Meaning} & \multirow[t]{2}{*}{FT} & Unconstrained & 0.774 & 0.413 & 0.702 & 0.630 & 0.682 & 0.691 & 0.705 & 0.580 & 0.647 \\
 &  &  & Constrained & 0.799 & 0.453 & 0.731 & 0.658 & 0.749 & 0.724 & 0.728 & 0.647 & 0.686 \\
 &  & \multirow[t]{2}{*}{PT} & Unconstrained & 0.819 & 0.532 & 0.767 & 0.693 & 0.694 & 0.758 & 0.778 & 0.632 & 0.709 \\
 &  &  & Constrained & 0.841 & 0.592 & 0.802 & 0.765 & 0.776 & 0.814 & 0.789 & 0.725 & 0.763 \\
 & \multirow[t]{4}{*}{Canonical} & \multirow[t]{2}{*}{FT} & Unconstrained & 0.799 & 0.486 & 0.781 & 0.647 & 0.724 & 0.732 & 0.725 & 0.619 & 0.689 \\
 &  &  & Constrained & 0.818 & 0.555 & 0.783 & 0.705 & 0.763 & 0.770 & 0.752 & 0.703 & 0.731 \\
 &  & \multirow[t]{2}{*}{PT} & Unconstrained & 0.839 & 0.544 & 0.777 & 0.729 & 0.770 & 0.791 & 0.789 & 0.702 & 0.743 \\
 &  &  & Constrained & 0.856 & 0.619 & 0.806 & \bfseries 0.779 & \bfseries 0.824 & 0.830 & 0.822 & \bfseries 0.793 & \bfseries 0.791 \\
\midrule
\multirow[t]{4}{*}{BART} & \multirow[t]{2}{*}{Meaning} & \multirow[t]{2}{*}{FT} & Unconstrained & 0.813 & 0.476 & 0.732 & 0.566 & 0.696 & 0.778 & 0.720 & 0.536 & 0.665 \\
 &  &  & Constrained & 0.834 & 0.499 & 0.750 & 0.619 & 0.739 & 0.796 & 0.774 & 0.620 & 0.704 \\
 & \multirow[t]{2}{*}{Canonical} & \multirow[t]{2}{*}{FT} & Unconstrained & 0.852 & 0.539 & 0.726 & 0.656 & 0.714 & 0.773 & 0.756 & 0.585 & 0.700 \\
 &  &  & Constrained & \bfseries 0.864 & 0.554 & 0.780 & 0.672 & 0.758 & 0.801 & 0.801 & 0.666 & 0.737 \\
\midrule
\multirow[t]{2}{*}{GPT-2} & Meaning & FT & Constrained & 0.760 & 0.479 & 0.736 & 0.571 & 0.645 & 0.699 & 0.660 & 0.606 & 0.644 \\
 & Canonical & FT & Constrained & 0.836 & 0.540 & 0.766 & 0.666 & 0.715 & 0.764 & 0.768 & 0.623 & 0.710 \\
\midrule
GPT-3 & Canonical & Context & Constrained & 0.859 & \bfseries 0.634 & 0.792 & 0.741 & 0.776 & 0.792 & 0.840 & 0.687 & 0.765 \\
\midrule
\multirow[t]{4}{*}{GPT-3*} & \multirow[t]{2}{*}{Meaning} & \multirow[t]{2}{*}{Context} & Unconstrained & 0.560 & 0.390 & 0.500 & 0.420 & 0.460 & 0.660 & 0.580 & 0.480 & 0.506 \\
 &  &  & Constrained & 0.680 & 0.530 & 0.680 & 0.580 & 0.630 & 0.750 & 0.780 & 0.630 & 0.657 \\
 & \multirow[t]{2}{*}{Canonical} & \multirow[t]{2}{*}{Context} & Unconstrained & 0.760 & 0.460 & 0.680 & 0.560 & 0.580 & 0.740 & 0.740 & 0.550 & 0.634 \\
 &  &  & Constrained & 0.800 & 0.620 & \bfseries 0.820 & 0.710 & 0.790 & \bfseries 0.840 & \bfseries 0.890 & 0.720 & 0.774 \\
\bottomrule
\end{tabular}

\caption{Results across all model size, target representation, tuning method, and decoding method for Overnight dataset. BART, GPT-2, and GPT-3 results results are included from \citet{shinConstrainedLanguageModels2021}}
\label{table:all-overnight-results}
\end{table*}

\end{document}